\documentclass{article}

\usepackage[nonatbib, final]{nips_2017}

\usepackage[utf8]{inputenc} % allow utf-8 input
\usepackage[T1]{fontenc}    % use 8-bit T1 fonts
\usepackage{hyperref}       % hyperlinks
\usepackage{url}            % simple URL typesetting
\usepackage{booktabs}       % professional-quality tables
\usepackage{amsfonts}       % blackboard math symbols
\usepackage{nicefrac}       % compact symbols for 1/2, etc.
\usepackage{microtype}      % microtypography

\usepackage{soul}
\usepackage{mathtools}
\usepackage{tabulary}
\usepackage{amssymb}
\usepackage{amsbsy}
\usepackage{mathtools}

\usepackage{graphics}
\usepackage{graphicx}
\usepackage{color}
\usepackage{verbatim}
\usepackage{algorithm}
\usepackage{algorithmic}
\usepackage{subcaption}

\usepackage{listings}
\usepackage{amsmath}
\usepackage{multirow}
\usepackage{multicol}

\def\ie{{\it i.e.,\ }}

\newcommand{\argmax}{\mathop{\rm arg~max}\limits}

\newcommand{\defeq}{\vcentcolon=}
\renewcommand{\r}{\mathbf{r}}

\newcommand{\vect}[1]{\boldsymbol{\mathbf{#1}}}

\newcommand{\vz}[0]{\vect{z}}

\renewcommand{\l}{^{(l)}}
\newcommand{\x}{\mathbf{x}}

\newcommand{\y}{\mathbf{y}}

\newcommand{\z}[1]{\vz^{(#1)}}
\newcommand{\hz}[1]{\hat{\vz}^{(#1)}}
\newcommand{\tz}[1]{\tilde{\vz}^{(#1)}}

\title{LLD Paper}  % for overleaf reference 
\title{Virtual Adversarial Ladder Networks for Semi-Supervised Learning}

\author{Saki Shinoda\textsuperscript{\rm1},\, Daniel E.\ Worrall\textsuperscript{\rm2} \& Gabriel J. Brostow\textsuperscript{\rm2} \\
Computer Science Department\\
University College London\\
United Kingdom \\
$^1$\texttt{saki.shinoda.16@ucl.ac.uk} \\
$^2$\texttt{\{d.worrall, g.brostow\}@cs.ucl.ac.uk} \\
}

\begin{document}
% \nipsfinalcopy is no longer used

\maketitle

\begin{abstract}
Semi-supervised learning (SSL) partially circumvents the high cost of labeling data by augmenting a small labeled dataset with a large and relatively cheap unlabeled dataset drawn from the same distribution. 
This paper offers a novel interpretation of two deep learning-based SSL approaches, ladder networks and virtual adversarial training (VAT), as applying distributional smoothing to their respective latent spaces. We propose a class of models that fuse these approaches. We achieve near-supervised accuracy with high consistency on the MNIST dataset using just 5 labels per class: our best model, ladder with layer--wise virtual adversarial noise (LVAN-LW), achieves $1.42\%\pm0.12$ average error rate on the MNIST test set, in comparison with $1.62\% \pm 0.65$ reported for the ladder network. On adversarial examples generated with $L_2$-normalized fast gradient method, LVAN-LW trained with 5 examples per class achieves average error rate $2.4 \% \pm 0.3$ compared to $68.6 \% \pm 6.5$ for the ladder network and $9.9 \% \pm 7.5$ for VAT.

\end{abstract}

\section{Introduction}
Ladder networks \cite{Valpola2014,Rasmus2015} and virtual adversarial training \cite{Miyato2016} are two seemingly unrelated deep learning methods that have been successfully applied to semi-supervised learning (SSL). Below we present the view that both methods share statistical information between labeled and unlabeled examples by smoothing the probability distributions over their respective latent spaces. With this interpretation, we propose a new class of deep models for SSL that apply spatially-varying, anisotropic smoothing to latent spaces in the direction of greatest curvature of the unsupervised loss function. 

In two models we investigate, we apply a virtual adversarial training cost in addition to ladder classification and denoising costs; in the other two models, we inject virtual adversarial noise into the encoder path. We train our models with 5, 10, or 100 labeled examples per class from the MNIST dataset, evaluating performance on both the standard test set and adversarial examples generated using the fast gradient method \cite{Goodfellow2014}. We found our models achieve state-of-the-art accuracy with high stability in the 5- or 10- labels per class setting on both normal and adversarial examples for MNIST. 

\section{Background}
In this section, we outline the ladder network \cite{Valpola2014,Rasmus2015} and Virtual Adversarial Training (VAT) \cite{Miyato2016}. We present the ladder network as representing data in a hierarchy of nested latent spaces. SSL is performed by smoothing the labeled and unlabeled data distributions over this hierarchy, thus sharing distributional information between labeled and unlabeled distributions in a coarse-to-fine regime. In VAT, there is no latent space hierarchy, but instead a particularly clever choice of smoothing operator.

\begin{figure}
\begin{subfigure}{0.5\textwidth}
\centering
\includegraphics[scale=0.22]{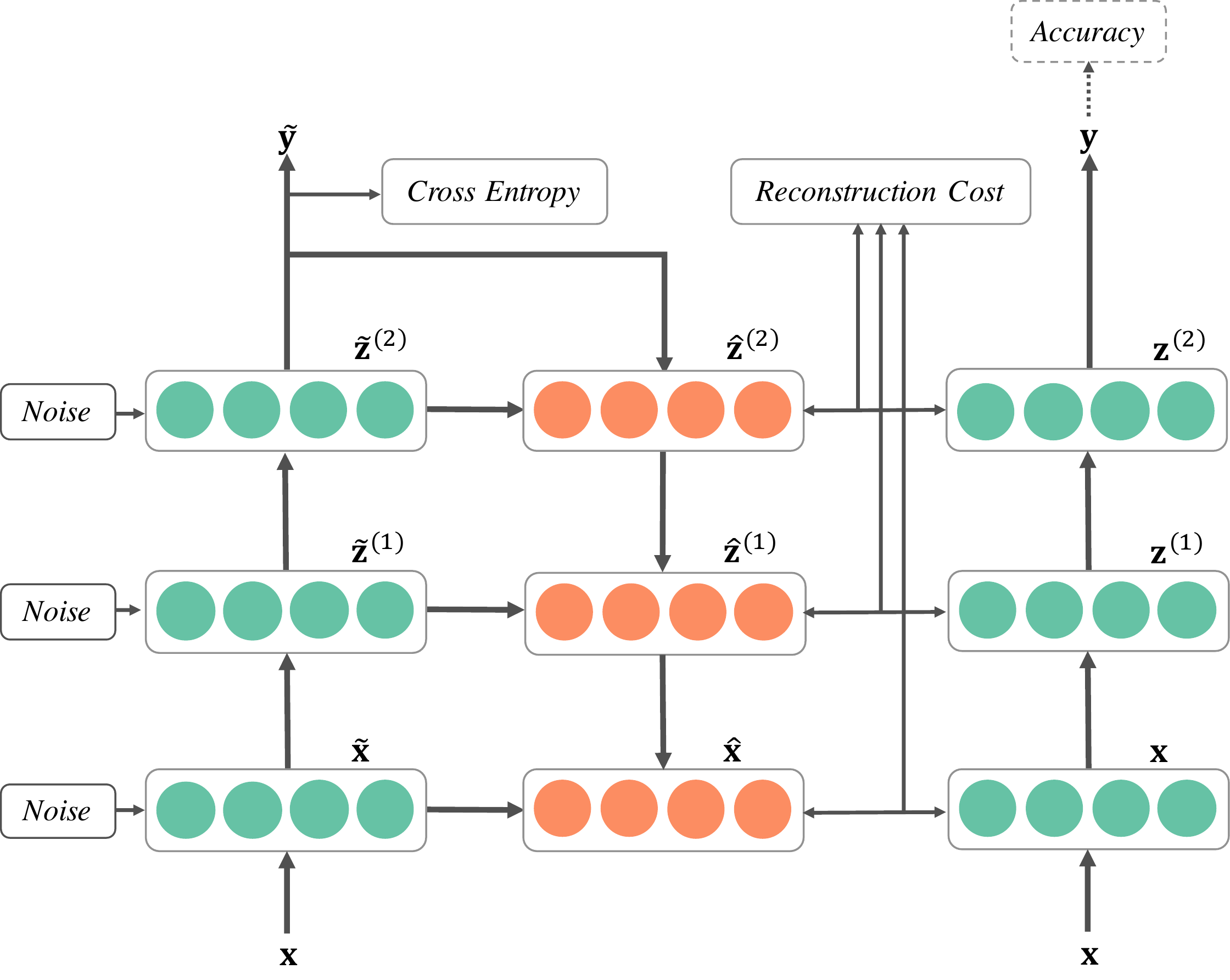}
\caption[Illustration of a ladder network architecture, showing the noisy encoder, decoder, and clean encoder.]{Illustration of a ladder network architecture.\\ \footnotesize \textbf{Left:} Noisy encoder with activations $\tz l$, additive Gaussian noise $\mathcal{N}(0,\sigma^2)$ at each layer, and outputs $\tilde {\vect y}$.\\ 
\textbf{Centre:} Decoder; input from layer above and corresponding layer in noisy encoder are combined by a denoising function $g\l(\cdot, \cdot)$ to form reconstructions $\hz l$.\\ 
\textbf{Right:} Clean encoder, weights shared with noisy encoder; activations $\vect z \l$ are denoising targets.}\label{fig:ladder}
\end{subfigure}
\hfill
\begin{subfigure}{0.45\textwidth}
\includegraphics[width=\textwidth]{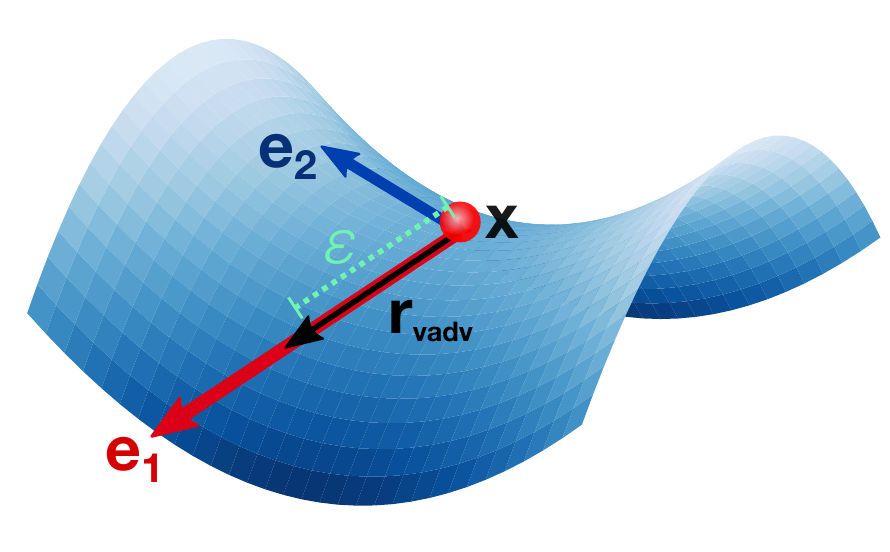}
\caption{Conceptual illustration of virtual adversarial perturbation $\vect r _{\rm vadv}$ on a surface representing the divergence $D \left[\Pr(y|\vect x, \vect \theta), \Pr(y | \vect x + \vect r, \vect \theta) \right]$. 
$\vect e_1$, $\vect e_2$ indicate eigenvectors of the Hessian $\nabla\nabla_r D |_{\vect r = \vect 0}$.
$\vect r_{\rm vadv}$ lies parallel to the dominant eigenvector $\vect e_1$ and has magnitude equal to the hyperparameter $\epsilon$. 
}
\end{subfigure}
\end{figure}

\paragraph{The Ladder Network}
For both supervised and unsupervised tasks, the ladder network \cite{Valpola2014,Rasmus2015} uses a single autoencoder-like architecture with added skip connections from encoder to decoder. For labeled examples, the encoder is used as a feed-forward classifier, and for the unsupervised task, the full architecture is used as a denoising autoencoder, with extra reconstruction costs on intermediate representations (see Figure \ref{fig:ladder}). For the denoising autoencoder, additive spherical Gaussian noise is applied to encoder activations $\mathbf{z}^{(\ell)}$, which we interpret as applying isotropic smoothing to the hierarchy of latent spaces modelled by the ladder network. We denoted the smoothed activations as $\tz \ell$. Mathematically, we have
\begin{align}
	\Pr (\{ \tz 1,..,\tz L \} | \x) &= \prod_{\ell=1}^L \int \mathcal{N} (\tz \ell; \z \ell, \sigma_{\text{Ladder}}^2 \vect I) \delta (\z \ell - \mathbf{y}^{(\ell-1)}) \, \mathrm{d} \tz \ell \label{eq:ladder_smoothing} \\
    \mathbf{y}^{(\ell)} &= f_{\ell}(\mathbf{W}^{(\ell)} \tz {\ell} + \mathbf{b}^{(\ell-1)}) 
\end{align}
where $\y^{(\ell-1)}$ is the output of the previous layer for $\ell > 0$ and $\y^{(0)} = \x$, $f_\ell$ is the nonlinearity, $\mathbf{W}^{(\ell)}$ is the weights, $\mathbf{b}^{(\ell)}$ is the bias. $ \mathcal{N} (\tz \ell; \z \ell, \sigma_{\text{Ladder}}^2 I)$ is a normal distribution over injected noise given mean $\z \ell$ and variance $\sigma_\text{Ladder}^2 \vect I$, $\vect I$ being the identity matrix.

The ladder network architecture is illustrated in Figure \ref{fig:ladder} for a ladder network with $L=2$ layers. %The encoder layers are numbered $l=0, 1, ..., L$ with $l=0$ corresponding to the inputs, and the decoder layers are numbered $l=L,L-1,...,0$ starting with $l=L$ the final output layer of the encoder. Activations at layer $l$ in the clean encoder are denoted $\vz^{(l)}$. The corresponding noisy activations in the corrupted encoder are denoted $\tz{l}$. The corresponding reconstructions, which are the activations of the decoder at $l$, are denoted $\hz{l}$. 
The ladder network is trained to simultaneously minimize a negative log-likelihood on labeled examples and a denoising reconstruction cost at each layer on unlabeled examples. Specifically, the reconstruction cost is a squared error between the decoder activation $\hz \ell$ and the noiseless encoder activation $\z \ell$. Forward passes at training time through the model can be seen as Monte Carlo sampling from equation \ref{eq:ladder_smoothing}.

\paragraph{Virtual Adversarial Training}
Virtual adversarial perturbations (VAP) were first presented in \cite{Miyato2016} extending adversarial perturbations from \cite{Goodfellow2014} to the case where there are no labels. Adversarial perturbations are computed as

\begin{equation}\label{eq:adv_ptb}
\r_{\rm adv} \defeq \argmax_{\r;\|\r\|\leq\epsilon} D\left[h(y), \Pr(y|\x + \r, \vect{\theta})\right], \end{equation} 
where $\epsilon$ is a parameter dictating the size of the perturbation; $h(y)$ is the target distribution, \ie a one-hot vector of the true labels; $\Pr(y|\x+\r,\vect{\theta})$ is the output probabilities of the model with parameters $\vect{\theta}$; and $D[p,q]$ is a statistical divergence between $P$ and $Q$ (\ie a positive definite functional, only equal to zero when $P=Q$), such as the Kullback--Leibler (KL) divergence \cite{MacKay2002}. VAPs are generated by approximating $h(y)$ with $\Pr(y|\x, {\vect \theta})$. The perturbation is defined as
\begin{equation}\label{eq:vadv_ptb}
\r_{\rm vadv} \defeq \argmax_{\r;\Vert \r \Vert _2 \leq \epsilon} D\left[\Pr(y|\x, {\vect \theta}), \Pr(y|\x + \r, \vect{\theta})\right], \end{equation}
In practice, as further detailed in \cite{Miyato2017}, the direction of $\vect r_{\rm vadv}$ is approximated via second-order Taylor expansion as the dominant eigenvector of the Hessian matrix of the divergence $D$ with respect to the perturbation $\vect{r}$; the eigenvector is computed with a finite difference power method \cite{Golub2000}.
% \begin{equation}
% \vect H (\x, \hat {\vect \theta}) \defeq \nabla \nabla_\r D \left . \left[ \Pr(y|\x, \hat{\vect \theta}), \Pr(y| \x + \r, \vect \theta) \right] \right |_{\r = \vect 0}
% \end{equation}
%This approximation for $\vect r _{\rm vadv}$ is parametrised by the finite difference constant $\xi$ and the number of power iterations $K$.
At training time, the virtual adversarial perturbed examples $\x + \r_{\rm vadv}$ are added to the training set and a virtual adversarial training loss is added to the current loss. The virtual adversarial training loss is defined as the average over all input data points of
\begin{equation}\label{eq:vadv_loss}
L_{\rm vadv}(\x,\vect{\theta}) \defeq D \left[\Pr(y|\x, {\vect \theta}), \Pr(y|\x+\r_{\rm vadv}, \vect \theta)\right].
\end{equation}
Minimization of this term can be seen as a smoothing operation, since it penalizes differences in $\Pr(y|\x, {\vect \theta})$, in its direction of greatest curvature. The range of smoothing and hence strength of regularization is controlled by the parameter $\epsilon$, the magnitude of the VAP. \cite{Miyato2017} found that fixing the coefficient of $L_{\rm vadv}$ relative to the supervised cost to 1 and tuning $\epsilon$ was sufficient to achieve good results, rather than tuning both. 

% In this paper, we use VAPs as both additive noise and to generate a cost term that explicitly smooths the output probability distribution. This latter approach is virtual adversarial training (VAT) from \cite{Miyato2016}. 

\section{Methods}

\begin{figure}[t]
\begin{subfigure}{0.24\textwidth}
\centering
\includegraphics[scale=0.22]{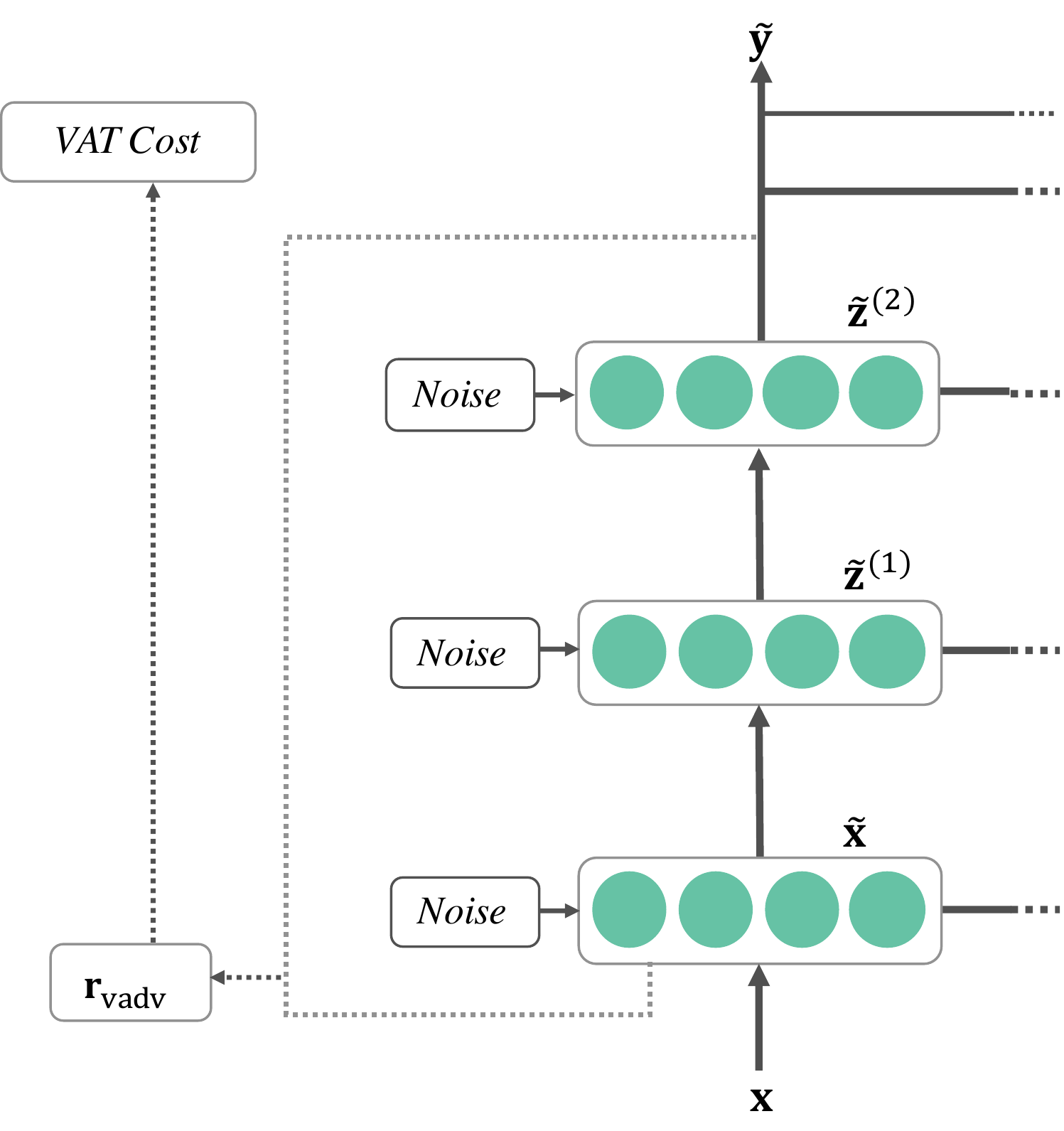}
\caption{LVAC}\label{fig:lvac}
\end{subfigure}
\hfill
\begin{subfigure}{0.24\textwidth}
\centering
\includegraphics[scale=0.22]{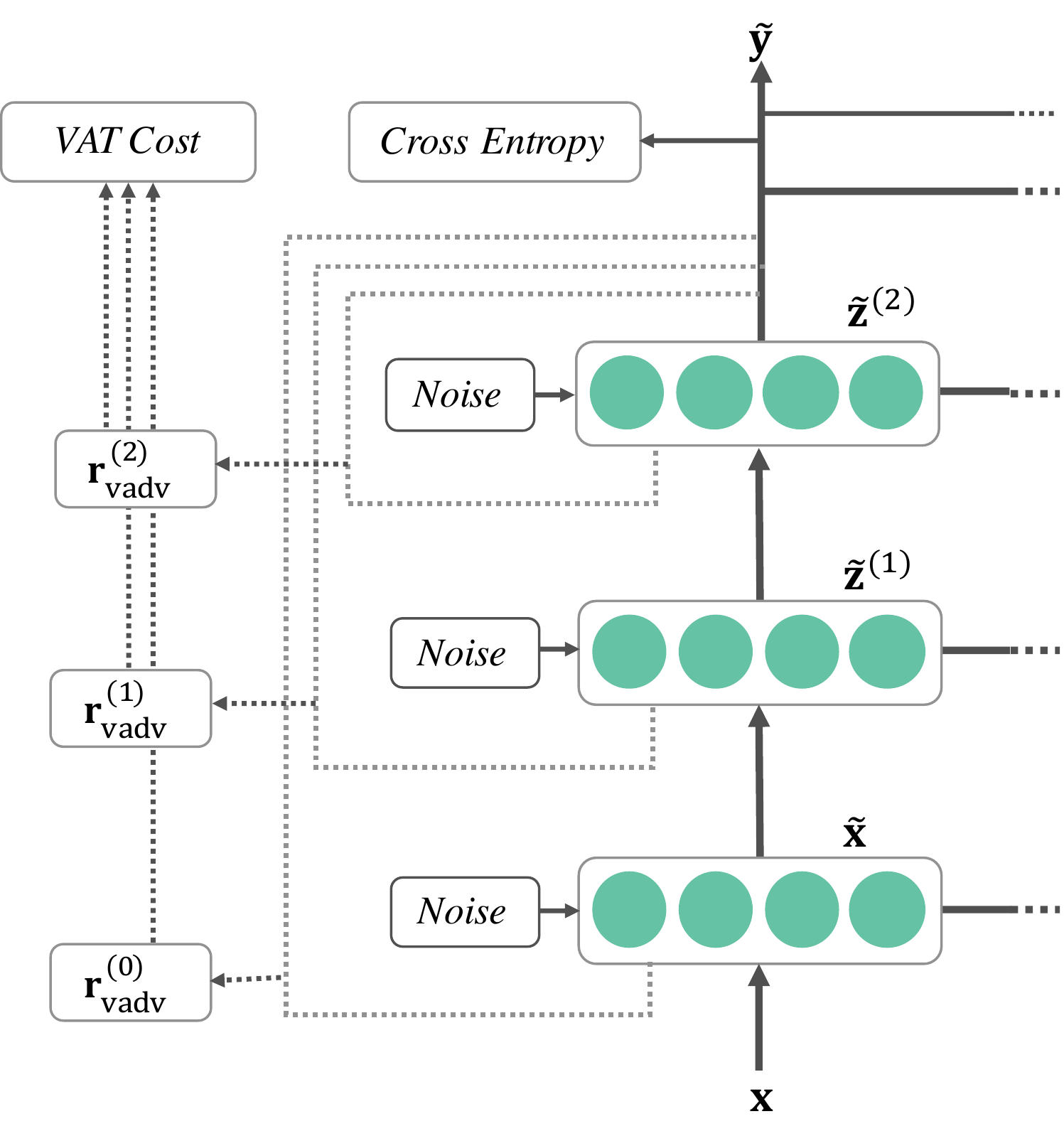}
\caption{LVAC-LW}\label{fig:lvaclw}
\end{subfigure}
\hfill
\begin{subfigure}{0.24\textwidth}
\centering
\includegraphics[scale=0.22]{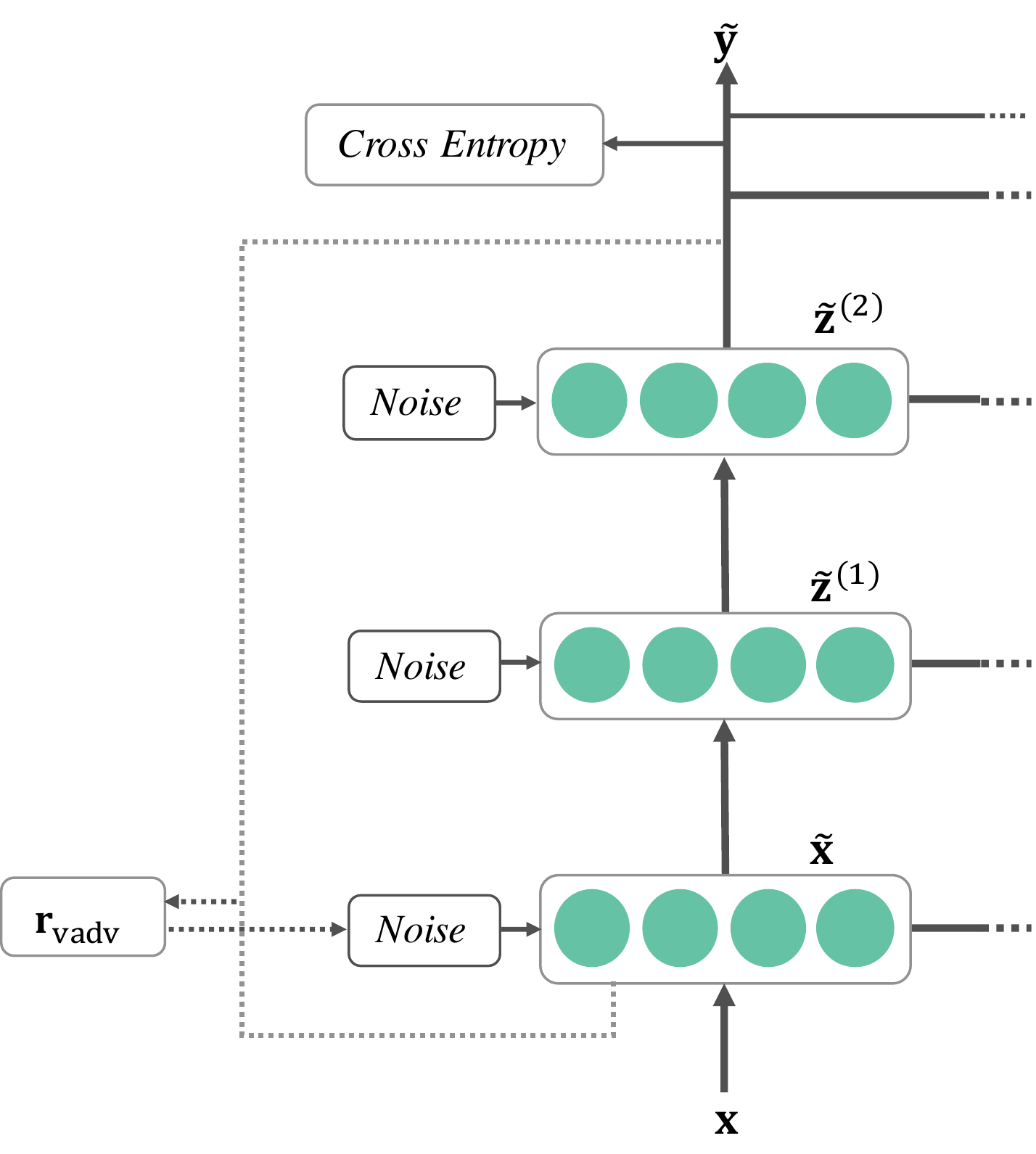}
\caption{LVAN}\label{fig:lvan}
\end{subfigure}
\hfill
\begin{subfigure}{0.24\textwidth}
\centering
\includegraphics[scale=0.22]{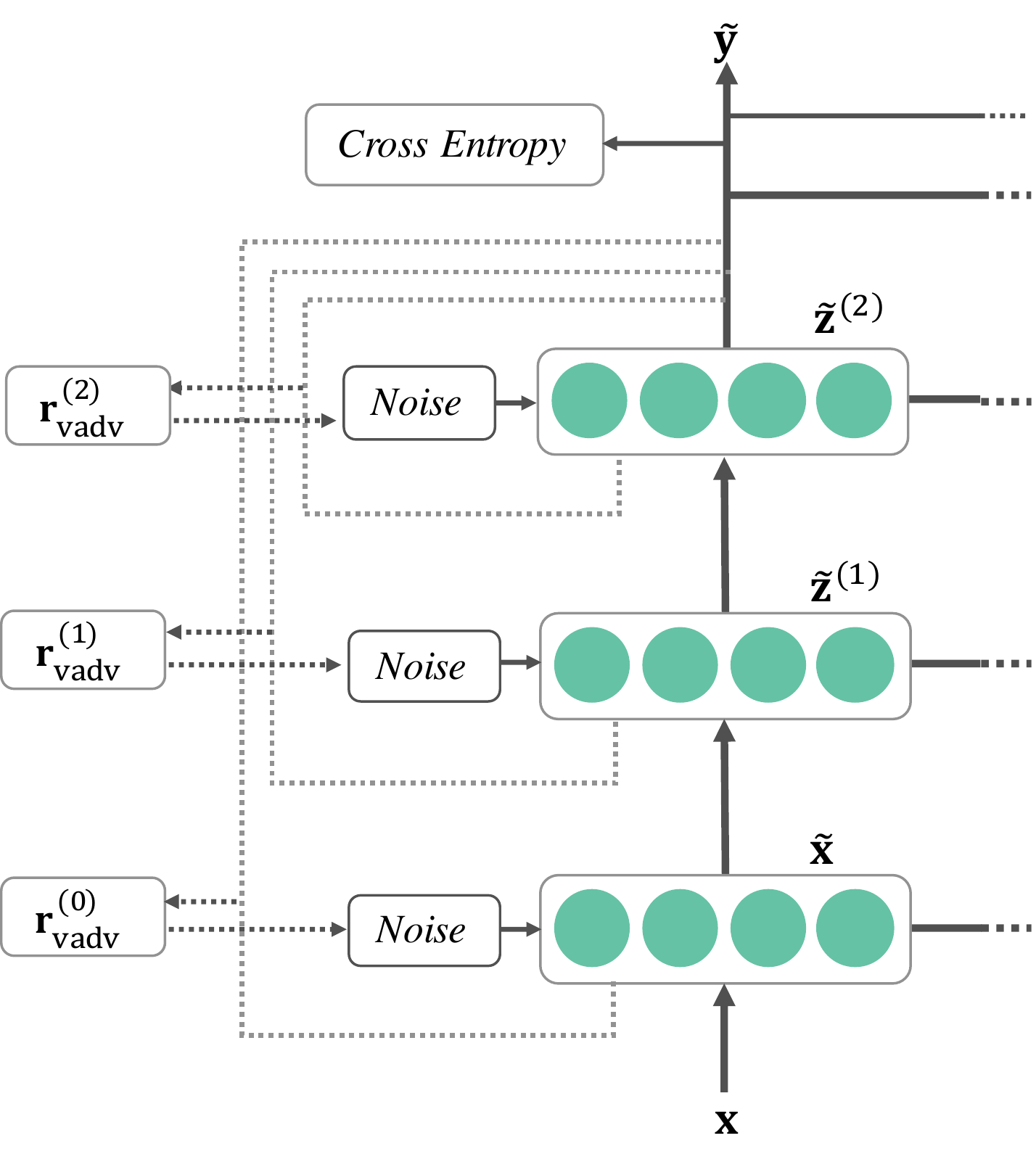}
\caption{LVAN-LW}\label{fig:lvanlw}
\end{subfigure}
\caption[Conceptual illustrations of our proposed models, (a) ladder with virtual adversarial cost (LVAC), (b) ladder with layer-wise virtual adversarial cost (LVAC-LW), (c) ladder with virtual adversarial noise (LVAN), (d) ladder with layer-wise virtual adversarial noise.]{Conceptual illustrations of our proposed models. All show corrupted encoder path only. Decoder and clean encoder (not shown) are identical to that of standard ladder in Figure \ref{fig:ladder}.}
\end{figure}

\paragraph{Ladder with virtual adversarial costs (LVAC and LVAC-LW)}
One approach for applying the anisotropic smoothing in output space of VAT to the ladder network is to add a virtual adversarial cost term (as in Equation \ref{eq:vadv_loss}) to the supervised cross-entropy cost and unsupervised activation reconstruction cost which is optimized to train the ladder network. In the most general formulation of VAT on a ladder, the loss term can be written 
\begin{align}
C_{\rm vadv} &= \sum_l \alpha \l D_{KL} \left[ 
\Pr(\tilde{y}|\tz l, \vect \theta),
\Pr(\tilde{y}|\tz l + \r_{\rm vadv}\l, \vect \theta)
\right],\label{eq:lw_cvadv}\\
\r_{\rm vadv} \l &= 
\argmax_{\r; \, \Vert \r \Vert _2 \leq \epsilon \l} D_{KL} \left[\Pr(\tilde y|\tz l, \vect \theta), \Pr(\tilde y|\tz l + \r, \vect{\theta})\right] \label{eq:lw_vadv}
\end{align}
where $\tz l$ is the activation in layer $l$ of the corrupted encoder path, with $\tz 0 = \tilde{\x}$. 
This gives the \emph{ladder with virtual adversarial costs, layer-wise (LVAC-LW)}, illustrated conceptually in \ref{fig:lvaclw}. Applying VAP to only the input images rather than the intermediate activations in the encoder gives us the \emph{ladder with virtual adversarial cost (LVAC)}, illustrated in Figure \ref{fig:lvac}.

\paragraph{Ladder with virtual adversarial noise (LVAN and LVAN-LW)} 
Alternatively, VAT smoothing can be applied to the ladder network by injecting virtual adversarial perturbations into the activations at each layer of the corrupted encoder in addition to isotropic Gaussian noise. VAP for each layer can be computed with the same form of Equation \ref{eq:lw_vadv} for any layer $l$. As with the models proposed above, the addition of noise can be on the input images only, giving the \emph{ladder with virtual adversarial noise (LVAN)} which is illustrated in Figure \ref{fig:lvan}. The alternative case of adding noise to each layer of the encoder, \emph{ladder with virtual adversarial noise, layer-wise (LVAN-LW)}, is shown in Figure \ref{fig:lvanlw}.

% \subsection{Additional hyperparameters}
% In LVAC-LW we compute a virtual adversarial perturbation for each layer, thus there is a tunable magnitude $\epsilon \l$ at each layer $l$. 
% As with LVAC-LW, there is a magnitude parameter $\epsilon \l$ for the virtual adversarial perturbations generated at each layer $l$, but as it has no explicit VAT cost term, like LVAN it has no associated $\alpha$ parameters.

\section{Experiments}

The structure of the classifier encoder for all of our models was a fully-connected network with layers of 1000, 500, 250, 250, 250, 10 units respectively. In all ladder implementations the decoder was symmetric to the encoder. All models were trained for 250 epochs using the Adam optimizer \cite{adam} with initial learning rate 0.002 and linear learning rate decay from 200 epochs. All models were trained with unlabeled batch size 100; labeled batch size was 50 for training with 50 labeled examples and 100 otherwise. VAP's were generated with $L_\infty$-norm. Hyperparameters were very roughly tuned using Bayesian optimization \cite{skopt}; values used are given in Appendix \ref{app:hyperparams}.

\paragraph{Low labeled data}
Table \ref{tbl:aer} shows average error rates on MNIST with 50, 100 and 1000 labeled examples for each of our proposed models and benchmark implementations of the ladder network and VAT. Mean and standard deviation for 50 labels is computed across ten training runs with different random seeds (fixed between models) for selecting labeled data and initializing weights; mean and standard deviations on 100 and 1000 labels are computed over five training runs. We expect high variability between training runs for very few labeled examples as performance depends significantly on the particular examples chosen. In this setting LVAN and LVAN-LW are notably highly stable in achieving very good performance.

%Our benchmark implementations of VAT and the ladder network perform slightly worse than reported results. Our LVAN-LW model 

\begin{table}[h]
\centering
\caption{Average Error Rate (\%) and standard error on MNIST}\label{tbl:aer}
\begin{tabular}{c cc cc cc}
\toprule
\bf Model & \multicolumn{2}{c}{\textbf{50 labels}} & \multicolumn{2}{c}{\textbf{100 labels}} & \multicolumn{2}{c}{\textbf{1000 labels}} \\
\midrule
VAT	(ours)	&5.38		& $\pm$ 2.92	& 2.14		& $\pm$ 0.64	& 1.11		& $\pm$ 0.05 \\
Ladder (ours) &1.86		& $\pm$ 0.43	& 1.45		& $\pm$ 0.36	& \bf{1.10}	& \bf{$\pm$ 0.05}  \\ 
LVAC		&2.42		& $\pm$ 1.05	& 1.65		& $\pm$ 0.12	& 1.28		& $\pm$ 0.07 \\
LVAC-LW		&4.08		& $\pm$ 3.55	& 1.39		& $\pm$ 0.06	& 1.11		& $\pm$ 0.12 \\
LVAN		&1.52		& $\pm$ 0.20	& 1.30		& $\pm$ 0.09	& 1.48		& $\pm$ 0.03 \\
LVAN-LW		&\bf{1.42}	&\bf{$\pm$ 0.12}	&\bf{1.25}	&\bf{$\pm$ 0.06}	& 1.51		& $\pm$ 0.06 \\
\bottomrule
\end{tabular}
\end{table}

\paragraph{Adversarial examples}
We tested the performance of our models on adversarial examples generated using the CleverHans implementation of the fast gradient method with $L_1$, $L_2$ and $L_\infty$ norms \cite{Goodfellow2014, Papernot2016}. 
For adversarial examples generated with $L_\infty$ norm, VAT outperformed the ladder network and all of our proposed models with 50, 100 and 1000 labeled examples. VAT and all of our models showed improvement in AER as the number of labeled examples increased, while the ladder network did not appear to improve substantially. On $L_1$ and $L_2$ adversarial examples, LVAN-LW followed closely by LVAN outperformed all models including VAT and ladder for 50, 100 labels, while LVAC and LVAC-LW performed better than LVAN, LVAN-LW, VAT and ladder for 1000 labels. The ladder network performed very poorly on 50 and 100 labels, while VAT AER for 1000 labels appears to have been limited by the relatively high AER on non-adversarial examples. The relative strength of the LVAC models on $L_\infty$ examples compared to $L_2$ or $L_1$ where the LVAN models performed better could be due to the relative strengths of regularization by VAT cost, which is based on $L_\infty$ perturbations, and the $L_2$ denoising cost. Full results are presented in Table \ref{tab:adv}.

\begin{table}[h]
\caption{Mean average error rate (\%) and standard error  on adversarial examples from MNIST}\label{tab:adv}
\begin{tabular}{cc *{6}{r@{ $\pm$ }l}}
\toprule
\bf Norm & \bf Labels & \multicolumn{2}{c}{VAT} & \multicolumn{2}{c}{Ladder} & \multicolumn{2}{c}{LVAC} & \multicolumn{2}{c}{LVAC-LW} & \multicolumn{2}{c}{LVAN} & \multicolumn{2}{c}{LVAN-LW} \\
\midrule
\multirow{3}{*}{$\boldsymbol L_\infty$} %
& 50   
&  \bf 22	 & \bf  5.3	        % VAT
&  54	     &  6              % Ladder
&  49	     &  2              % LVAC
&  34	     &  6              % LVAC-LW
&  59	     &  5              % LVAN
&  56	     &  3              % LVAN-LW
\\ 
& 100  
&  \bf 16	 & \bf  1	        % VAT
&  58	     &  2              % Ladder
&  48	     &  1              % LVAC
&  31	     &  2              % LVAC-LW
&  60	     &  3              % LVAN
&  60	     &  2              % LVAN-LW 
\\
& 1000 
&  \bf 10.6	 & \bf  0.4	        % VAT
&  53	     &  2              % Ladder
&  36	     &  1              % LVAC    
&  27	     &  3              % LVAC-LW 
&  40	     &  1              % LVAN    
&  43	     &  2              % LVAN-LW
\\ 
\midrule[0.5\lightrulewidth]
\multirow{3}{*}{$ \boldsymbol L_2$} %
&  50 
&  10	        &  8	        % VAT
&  26           &  5	        % Ladder
&  3	        &  2	        % LVAC
&  5	        &  4	        % LVAC-LW
&  1.8	        &  0.4	        % LVAN
&  \bf 1.6	    & \bf  0.2	    % LVAN-LW
\\
& 100 
&  3	        &  2	        % VAT
&  1.6	        &  0.4	        % Ladder
&  1.7	        &  0.1	        % LVAC
&  1.4	        &  0.1	        % LVAC-LW
&  1.4	        &  0.1	        % LVAN
&  \bf 1.3 	    & \bf  0.1	    % LVAN-LW
\\
& 1000
&  2.4	        &  0.2	        % VAT
&  1.5	        &  0.1	        % Ladder
&  1.4	        &  0.1	        % LVAC    
&  \bf 1.2	    & \bf  0.2	    % LVAC-LW    
&  1.6	        &  0.1	        % LVAN    
&  1.7	        &  0.1	        % LVAN-LW   
\\ 
\midrule[0.5\lightrulewidth]
\multirow{3}{*}{$\boldsymbol L_1$} %
& 50 
&  10	     &  8	            % VAT
&  69	     &  7	            % Ladder
&  3	     &  2	            % LVAC
&  4	     &  4	            % LVAC-LW
&  2.5	     &  0.4	            % LVAN
&  \bf 2.4	 & \bf  0.3	        % LVAN-LW
\\
& 100
&  4	     &  2	            % VAT
&  2.3	     &  0.4	            % Ladder
&  2.2	     &  0.1	            % LVAC
&  \bf 1.9	 &  \bf 0.1	        % LVAC-LW
&  2.0	     &  0.2	            % LVAN
&  \bf 1.9	 & \bf  0.2	        % LVAN-LW
\\
& 1000                           
& 2.6	     &  0.2	            % VAT    
& 2.3	     &  0.2	            % Ladder    
& \bf 1.8	 &  \bf 0.2	        % LVAC    
& 1.6	     &  0.1	            % LVAC-LW    
& 2.2	     &  0.1	            % LVAN    
& 2.3	     &  0.2	            % LVAN-LW    
\\
\bottomrule
\end{tabular}      
\end{table}

\paragraph{Introspection: measuring smoothness}
\begin{figure}[t]
\centering
\caption{Comparison of average error rate and virtual adversarial cost (3 runs of 150 epochs each).}\label{fig:rvadv}
\includegraphics[width=\textwidth]{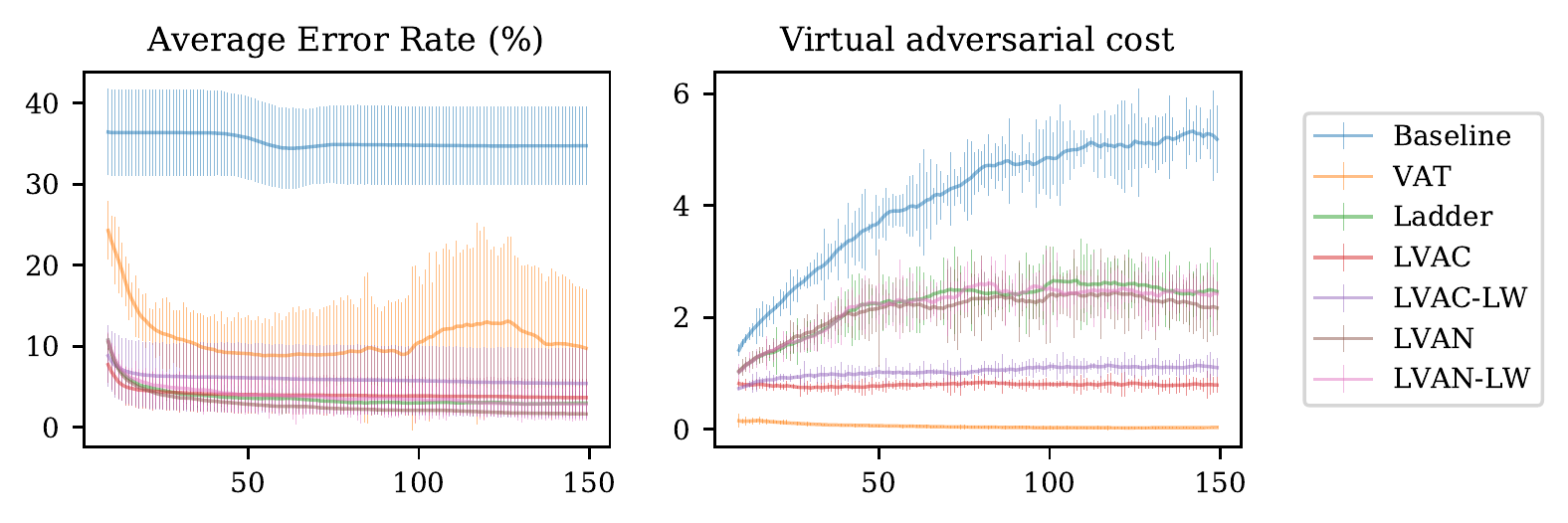}
\end{figure}
The virtual adversarial loss given in Equation \ref{eq:vadv_loss} is a measure of \emph{lack of} local smoothness of the output distribution with respect to the input images. We expect that the ladder and LVAN models, though they do not explicitly minimize this cost, still perform smoothing that should be reflected in this metric.
This cost was computed with $\epsilon=5.0$ for all models over 150 epochs of training (Figure \ref{fig:rvadv}). As expected, VAT, which directly minimizes this cost, is most smooth by this metric. The benchmark ladder network is significantly smoother than the fully supervised baseline despite not explicitly minimising the virtual adversarial cost. We measure LVAN and LVAC to be smoother than LVAN-LW and LVAC-LW respectively. This suggests smoothing with respect to the input image, which this metric measures, is traded off in the layer-wise models with smoothing in the intermediate latent spaces.

% Another measurement that easily arises from the machinery of generating VAP's is the spectral radius of the Hessian matrix.  The spectral radius can be computed using the power method and Rayleigh quotient \cite{Golub2000}. 
% % The power method iterates the computation 
% % \begin{equation}
% % \vect q_{k+1} \gets \frac{\vect H \vect q_k}{ \left( \vect H \vect q_k \right)^\top \left(\vect H \vect q_k \right)}
% % \end{equation}
% % for $\vect q$ to converge to the dominant eigenvector. The dominant eigenvalue can be approximated by the Rayleigh quotient:
% % \begin{equation}
% % \frac{ \vect q _{k+1} \cdot \vect q _{k}}{\vect q _{k} ^\top \vect q_k}
% % \end{equation}

% We take the absolute of this value as the spectral radius. \hl{Maybe best not to include the spectral radius since instabilities cause NaNs which have been forward filled for plotting purposes. }

\section{Discussion and Conclusions}
In this work, we conducted an analysis of the ladder network from \cite{Rasmus2015} and virtual adversarial training (VAT) from \cite{Miyato2016, Miyato2017} for semi-supervised learning and proposed four variants of a model applying virtual adversarial training to the ladder network: ladder with virtual adversarial cost (LVAC), ladder with layer-wise virtual adversarial cost (LVAC-LW), ladder with virtual adversarial noise (LVAN), and ladder with layer-wise virtual adversarial noise (LVAN-LW). 

Based on the manifold and cluster assumptions of semi-supervised learning \cite{Chapelle2006}, we hypothesised that virtual adversarial training could improve the classification accuracy of the ladder network trained in a semi-supervised context. We tested this hypothesis on the MNIST dataset \cite{Lecun1998}, by training models on training sets consisting of 50, 100 or 1000 labeled examples augmented by the full 60,000 images in the MNIST training set as unlabeled examples. We measured performance as error rate on the held-out test set of 10,000 examples.

We found that our models, most significantly LVAN-LW, improved on the performance of the ladder for 50 labels and 100 labels, achieving state-of-the-art error rates. For 1000 labels, both VAT and ladder baselines outperformed our models. This leads us to believe that the additional regularization provided by VAP's to the ladder network are useful only when the task is sufficiently challenging, suggesting that we should test our models on more complex datasets such as SVHN or CIFAR-10.

Additionally we found that our models performed better than the ladder network on adversarial examples. VAT outperformed our models for $L_\infty$ adversarial examples, but our models, again especially the LVAN-LW model, achieved best performance for the few-label cases (50 and 100 labels) on $L_1$ and $L_2$-normalized adversarial examples.

Our best-performing models overall were based on adding virtual adversarial noise to the corrupted encoder path of the ladder (LVAN and LVAN-LW). These have additional advantages over the LVAC models proposed: they are faster, as they require fewer passes through the network, and produce more stable, consistent results.

A natural extension of our work would be to extend our interpretation to deep SSL methods which have been more recently introduced such as temporal ensembling \cite{Laine2016}, random data augmentation \cite{Sajjadi2016}, and the Mean Teacher method \cite{Tarvainen2017}.

\section*{Acknowledgements}
This work was carried out by Saki Shinoda while at UCL and completed while at Prediction Machines Pte Ltd. Daniel Worrall is funded by Fight For Sight, UK.

\section*{Code}
Code available at \url{https://github.com/sakishinoda/tf-ssl/}

\bibliographystyle{abbrv}
\bibliography{nips_2017}

\begin{thebibliography}{10}

\bibitem{Chapelle2006}
O.~Chapelle, B.~Sch{\"{o}}lkopf, and A.~Zien.
\newblock {\em {Semi-Supervised Learning}}.
\newblock MIT Press, Cambridge, Massachusetts, 2006.

\bibitem{Golub2000}
G.~H. Golub and H.~A. van~der Vorst.
\newblock Eigenvalue computation in the 20th century.
\newblock {\em J. Comput. Appl. Math.}, 123(1-2):35--65, Nov. 2000.

\bibitem{Goodfellow2014}
I.~J. {Goodfellow}, J.~{Shlens}, and C.~{Szegedy}.
\newblock {Explaining and Harnessing Adversarial Examples}.
\newblock {\em arXiv preprint arXiv:1412.6572}, Dec. 2014.
\newblock Presented at the 3rd International Conference on Learning
  Representations (San Diego, CA, USA, 7--9 May 2015).

\bibitem{adam}
D.~P. {Kingma} and J.~{Ba}.
\newblock {Adam: A Method for Stochastic Optimization}.
\newblock {\em arXiv preprint arXiv:1412.6980}, Dec. 2014.
\newblock Presented at the 3rd International Conference on Learning
  Representations (San Diego, CA, USA, 7--9 May 2015).

\bibitem{Laine2016}
S.~{Laine} and T.~{Aila}.
\newblock {Temporal Ensembling for Semi-Supervised Learning}.
\newblock {\em ArXiv e-prints}, Oct. 2016.
\newblock Presented at the 5th International Conference on Learning
  Representations (Toulon, FR, 24--26 April 2017).

\bibitem{Lecun1998}
Y.~Lecun, L.~Bottou, Y.~Bengio, and P.~Haffner.
\newblock Gradient-based learning applied to document recognition.
\newblock {\em Proceedings of the IEEE}, 86(11):2278--2324, Nov 1998.

\bibitem{MacKay2002}
D.~J.~C. MacKay.
\newblock {\em Information Theory, Inference \& Learning Algorithms}.
\newblock Cambridge University Press, New York, NY, USA, 2002.

\bibitem{Miyato2017}
T.~{Miyato}, S.-i. {Maeda}, M.~{Koyama}, and S.~{Ishii}.
\newblock {Virtual Adversarial Training: a Regularization Method for Supervised
  and Semi-supervised Learning}.
\newblock {\em arXiv preprint arXiv:1704.03976}, Apr. 2017.

\bibitem{Miyato2016}
T.~{Miyato}, S.-i. {Maeda}, M.~{Koyama}, K.~{Nakae}, and S.~{Ishii}.
\newblock {Distributional Smoothing with Virtual Adversarial Training}.
\newblock {\em arXiv preprint arXiv:1507.00677}, July 2015.
\newblock Presented at the 4th International Conference on Learning
  Representations (San Juan, PR, USA, 2--4 May 2016).

\bibitem{Papernot2016}
N.~Papernot, I.~Goodfellow, R.~Sheatsley, R.~Feinman, and P.~McDaniel.
\newblock cleverhans v1.0.0: an adversarial machine learning library.
\newblock {\em arXiv preprint arXiv:1610.00768}, 2016.

\bibitem{Rasmus2015}
A.~Rasmus, H.~Valpola, M.~Honkala, M.~Berglund, and T.~Raiko.
\newblock Semi-supervised learning with ladder networks.
\newblock In {\em Proceedings of the 28th International Conference on Neural
  Information Processing Systems}, NIPS'15, pages 3546--3554, Cambridge, MA,
  USA, 2015. MIT Press.

\bibitem{Sajjadi2016}
M.~Sajjadi, M.~Javanmardi, and T.~Tasdizen.
\newblock Regularization with stochastic transformations and perturbations for
  deep semi-supervised learning.
\newblock In {\em Advances in Neural Information Processing Systems}, pages
  1163--1171, 2016.

\bibitem{skopt}
{scikit-optimize contributors}.
\newblock {Scikit-Optimize}: Sequential model-based optimization with a
  `scipy.optimize` interface, 2017.
\newblock Software used under BSD license.

\bibitem{Tarvainen2017}
A.~Tarvainen and H.~Valpola.
\newblock Weight-averaged consistency targets improve semi-supervised deep
  learning results.
\newblock {\em arXiv preprint arXiv:1703.01780}, 2017.
\newblock Accepted as a conference paper at 2017 conference on Neural
  Information Processing Systems.

\bibitem{Valpola2014}
H.~{Valpola}.
\newblock {From neural PCA to deep unsupervised learning}.
\newblock {\em arXiv preprint arXiv:1411.7783}, Nov. 2014.

\end{thebibliography}

\appendix
\section{Hyperparameters}\label{app:hyperparams}

\begin{table}[H]
\centering
\begin{tabular}{cc ccc ccc}
\toprule
\bf Model & \textbf{Labels}	& $\lambda^{(0)}$ & $\lambda^{(1)}$ & $\lambda^{(\geq 2)}$ & $\epsilon^{(0)}$ & $\epsilon^{(1)}$ & $\epsilon^{(\geq 2)}$\\
\midrule
\multirow{3}{*}{LVAC} 
& 50	    &	1504	&	16.15	&	0.0381	&	0.0733	&   -       &   -                       \\				
& 100	    &	1966	&	14.20	&	0.1563	&	0.0731	&   -       &   -                       \\				
& 1000	    &	3883	&	12.35	&	0.0539	&	2.5206	&   -       &   -                       \\				
\midrule													
\multirow{3}{*}{LVAC-LW} 
& 50	    &	1000	&	10.00	&	0.1000	&	1.0000	&	0.1000	&	1.00 $\times$10$^{-3}$	\\
& 100	    &	1966	&	14.20	&	0.1563	&	0.0731	&	0.4822	&	1.402 $\times$10$^{-3}$	\\
& 1000	    &	3883	&	12.35	&	0.0539	&	2.5206	&	0.0143	&	6.002 $\times$10$^{-4}$	\\
\midrule
\multirow{3}{*}{LVAN}
& 50	    &	1504	&	16.15	&	0.0381	&	0.0733	&   -       &   -                       \\				
& 100	    &	1966	&	14.20	&	0.1563	&	0.0731	&   -       &   -                       \\								
& 1000	    &	3883	&	12.35	&	0.0539	&	2.5206	&   -       &   -                       \\				
\midrule										
\multirow{3}{*}{LVAN-LW}
& 50	    &	1504	&	16.15	&	0.0381	&	0.0733	&	0.3897	&	8.372 $\times$10$^{-2}$	\\
& 100	    &	1966	&	14.20	&	0.1563	&	0.0731	&	0.4822	&	1.402 $\times$10$^{-3}$	\\
& 1000	    &	3883	&	12.35	&	0.0539	&	2.5206	&	0.0143	&	6.002 $\times$10$^{-4}$	\\
\midrule								
\multirow{3}{*}{Ladder}
& 50	    &	1504	&	16.15	&	0.0381	&   -       &   -       &   -                       \\						
& 100	    &	1966	&	14.20	&	0.1563	&   -       &   -       &   -                       \\						
& 1000  	&	3883	&	12.35	&	0.0539	&   -       &   -       &   -                       \\						
\midrule
\multirow{3}{*}{VAT}
& 50	    &  -        &   -       &   -       &   5.0	    &   -       &   -                       \\						
& 100	    &  -        &   -       &   -       &   5.0	    &   -       &   -                       \\						
& 1000  	&  -        &   -       &   -       &   2.5	    &   -       &   -                       \\						
\bottomrule
\end{tabular}
\end{table}

\end{document}